\pdfoutput=1

\documentclass[11pt]{article}

\usepackage{acl}
\usepackage{amsmath}
\usepackage{times}
\usepackage{latexsym}
\usepackage{booktabs} 
\usepackage{float}
\usepackage{multirow}
\usepackage{soul}
\usepackage[T1]{fontenc}

\usepackage[utf8]{inputenc}

\usepackage{microtype}

\usepackage{inconsolata}

\usepackage{graphicx}
\usepackage{pifont}

\title{Relation Extraction Across Entire Books to Reconstruct Community Networks: The \textsc{AffilKG} Datasets}

\author{\quad Erica Cai \quad Sean McQuade \quad Kevin Young \quad Brendan O'Connor\\
  University of Massachusetts Amherst \\
  \texttt{\{ecai,spmcquade,keviny,brenocon\}@umass.edu} \\}

\begin{document}
\maketitle
\begin{abstract}

When knowledge graphs (KGs) are automatically extracted from text, are they accurate enough for downstream analysis? Unfortunately, current annotated datasets can not be used to evaluate this question, since their KGs are highly disconnected, too small, or overly complex.
To address this gap, we introduce \textsc{AffilKG}\footnote{\url{https://doi.org/10.5281/zenodo.15427977}}, which is a collection of six datasets that are the first to pair complete book scans with large, labeled knowledge graphs. Each dataset features affiliation graphs, which are simple KGs that capture \textsc{Member} relationships between \textsc{Person} and \textsc{Organization} entities---useful in studies of migration, community interactions, and other social phenomena. In addition, three  datasets include expanded KGs with a wider variety of relation types. Our preliminary experiments demonstrate significant variability in model performance across datasets, underscoring \textsc{AffilKG}'s ability to enable two critical advances: (1) benchmarking how extraction errors propagate to graph-level analyses (e.g., community structure), and (2) validating KG extraction methods for real-world social science research.  

\end{abstract}

\begin{figure*}
    \centering
    \includegraphics[scale=0.39]{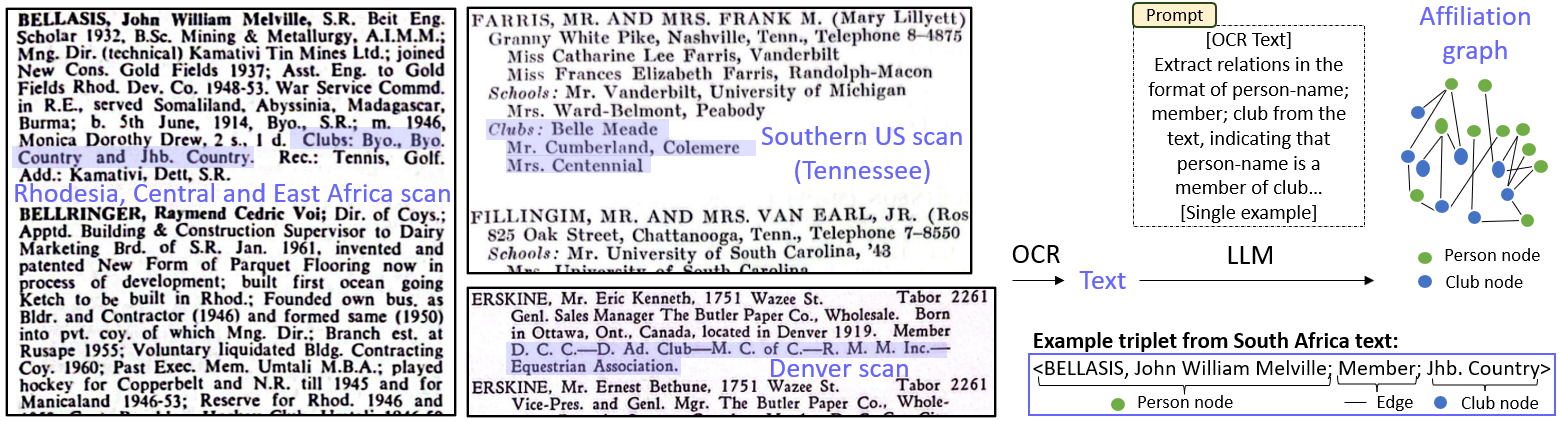}
    \caption{Pipeline of converting scans of historical text to affiliation graphs, where highlighted text indicates clubs.
    }
    \vspace{-1.1em}
    \label{fig:pipeline}
\end{figure*}

\section{Introduction}

Extracting social networks from text is of great interest to social scientists for understanding how elite communities interact \cite{obrien2025gender,Broad1996social} and how network structures change across time \cite{Wellman1996social,kossinets2006empirical}. Historically, many networks have been manually curated with high precision \cite{edwards2020one}, such as financial acquisitions from news articles \cite{GUGLER2003625,clougherty2014cross} or interpersonal relationships in literary sagas \cite{carron2013sagas}. However, manual annotation is labor-intensive and difficult to scale, while automated approaches often lack rigorous evaluation \cite{DRURY2022social}, raising concerns about their reliability for downstream analyses.

In parallel, hundreds of papers on relation extraction (RE), a.k.a.\ knowledge graph (KG) extraction,
aim to automate the construction of these networks as KGs, which represent information using triplets (e.g., $\langle$\textsc{Bob}; \textsc{member}; \textsc{IEEE}$\rangle$), each of which encodes a relation (\textsc{member}) between two entities (\textsc{Bob}, \textsc{IEEE}) that is expressed in text.
In a KG, entities are represented as nodes and relations as edges (Fig.~\ref{fig:pipeline}) \cite{Ding2021PrototypicalRL,nadgeri-etal-2021-kgpool,xu-barbosa-2019-connecting,trisedya-etal-2019-neural,han-etal-2021-exploring}. Despite prior poor large language model (LLM) performance for RE \cite{jimenez-gutierrez-etal-2022-thinking}, recent works show increasingly better performance with limited training data \cite{wadhwa-etal-2023-revisiting,wan-etal-2023-gpt,zhou2024leap,zhang-soh-2024-extract,tao2024graphicalreasoningllmbasedsemiopen,xue-etal-2024-autore,Zhang2024-cw,rajpoot-parikh-2023-gpt}. However, a fundamental question remains: Do widely used evaluations
for RE and KG extraction \cite{zhao2024comprehensive,DETROJA2023survey} reliably indicate performance on higher-level, practical graph analyses, such as identifying key individuals in a social network \cite{cai2024from}?

Most existing NLP datasets for RE make it difficult to assess how KG extraction errors affect downstream graph analyses. Many datasets label triplets that fail to define connected KGs because the underlying text consists of unrelated sentences and disjoint entities \cite{zhang-etal-2017-position, han-etal-2018-fewrel, Riedel2010ModelingRA, chen-etal-2021-h, gardent-etal-2017-creating, huguet-cabot-etal-2023-red}. Others involve highly complex KGs with hundreds of relation types or label KGs in short documents, resulting in small, disconnected graphs \cite{yao-etal-2019-docred}. Even when datasets yield connected KGs, low extraction performance (often below 50 F1 points) limits meaningful insights into downstream analyses that are relevant for real world applications \cite{luan-etal-2018-multi, jain-etal-2020-scirex}.

To address these challenges, we introduce \textsc{AffilKG}, the first collection of datasets that pair full scanned books---along with their text extracted via optical character recognition (OCR)---with large, labeled knowledge graphs. Each of the six datasets in \textsc{AffilKG} provides an affiliation graph, which is a bipartite KG where edges denote \textsc{Member} relationships between \textsc{Person} and \textsc{Club} entities. These graphs are particularly valuable for studying community cohesion, interaction dynamics \cite{PREMSANKAR2015exploratory, tisch2023top}, and broader topics such as social stratification and network evolution \cite{corradi2024, bichler2015white, young2015diffusion, grund2015ethnic}. Three datasets are further expanded to include triplets of 4 to 11 additional relation types, such as \textsc{College} and \textsc{Yacht Name}, enabling richer relational analyses.

The texts in \textsc{AffilKG} are unique archival repositories of information on elite families in specific locations. Known as `social registers', these texts document elite individuals from diverse geographic regions and time periods, including the Southern US \cite{whitmarsh1951}, Denver \cite{1931denver}, Rhodesia \cite{1963rhodesia}, Boston \cite{1903BostonSocialRegister}, Miami \cite{miami1965}, and New York City \cite{1892NewYorkSocialRegister}. Social register texts served to connect elite families with one another in a common directory of an area, typically listing family members, addresses, club memberships and educational backgrounds. As a historical text, a given social register thus provides a comprehensive record of a local social network, making it ideal for evaluating RE methods their downstream impact on graph analyses. Affiliation graph triplets were generated either through manual annotation or by applying regular expressions and rule-based algorithms, which were rigorously validated against manually constructed ground truth.

\textsc{AffilKG} addresses critical limitations in existing datasets and provides a foundation for studying how RE errors impact downstream graph analyses. We present preliminary results using both widely used NLP evaluation metrics and higher-level analyses relevant to real-world applications. Our findings indicate that, while extraction performance can be high enough to yield meaningful insights, there is also substantial variation across settings. This suggests that further investigation into the propagation of RE errors within graph analyses could uncover important patterns and provide actionable knowledge for practical use. 

\section{\textsc{AffilKG} Dataset Construction}
\label{sec:datasets}

\textsc{AffilKG} contains six datasets pairing scanned historical social register books with large, labeled knowledge graphs.
Each individual dataset includes an affiliation graph, and three provide expanded KGs with additional relation types. All datasets include high-quality scans, OCR text, and graph annotations, enabling robust evaluation of relation extraction and higher-level graph analyses (Fig.~\ref{fig:pipeline}).

\begin{table}
{\footnotesize\begin{tabular}{lccc}
 Loc./Year & \begin{tabular}{l} 
\# \\
pages
\end{tabular} & \begin{tabular}{l} 
\#~nodes \\
(person / org)
\end{tabular} & \begin{tabular}{l} 
\# \\
edges\end{tabular} \\
\midrule 
NYC/1892 & 299 &  5553 / 178 & 14061 \\
Boston/1903 & 150 &  2534 / 188 & 6228\\
Denver/1931 & 34 & 358 / 248 & 1352 \\
S. US./1951 & & & \\
{\fontsize{8pt}{8pt}\selectfont \hspace{0.6em} All (13 states)} & {\fontsize{7.6pt}{8pt}\selectfont 826}& {\fontsize{7.6pt}{8pt}\selectfont 6188 / 3107  } & {\fontsize{7.6pt}{8pt}\selectfont 13262} \\ 
{\fontsize{8pt}{8pt}\selectfont \hspace{0.6em} Median: Tenn.} & {\fontsize{7.6pt}{8pt}\selectfont 47}& {\fontsize{7.6pt}{8pt}\selectfont 399 / 174  } & {\fontsize{7.6pt}{8pt}\selectfont 872} \\ 
Rhod./1963 & 169&  898 / 767 & 2367 \\
Miami/1965 & 503 &  3775 / 1471 & 11830 \\

\hline
\end{tabular}}

\caption{Details about \textit{Denver}, \textit{Southern US} (13 states aggregated), \textit{Tennessee} (state with the median \# of person nodes), \textit{Rhodesia}, \textit{Miami}, \textit{Boston}, and \textit{New York City} text/affiliation graph pairs.}
\label{tab:affil-graph-details}
\end{table}

\begin{table}
{\footnotesize
\begin{tabular}{lccc}
Loc./Year & \begin{tabular}{l} 
\# 
nodes \\
(person / other)
\end{tabular} & \begin{tabular}{l} 
\# \\
edges\end{tabular} & \begin{tabular}{l} 
\# 
rel \\ types
\end{tabular} \\
\midrule
NYC/1892 & 13054 / 5182 & 29263 & 4 \\
Boston/1903 & 6748 / 4619 & 18945 & 5 \\
Miami/1965 & 6340 / 8065 & 30845 & 11 \\
\hline
\end{tabular}
}
\caption{Details about \textit{Miami}, \textit{Boston}, and \textit{New York City} expanded KGs.}
\label{tab:expanded-kg-details}
\end{table}

\vspace{0.3em}
\noindent\textbf{Texts.} The books in \textsc{AffilKG} are archival social registers, which serve as historical "who's who" directories that catalog elite individuals and their affiliations within a locality at a specific time, and have been used by social scientists aiming to represent elite populations and their networks 
\cite{obrien2024family,Broad2020social,Baltzell1958philadelphia}. These sources connect prominent figures to clubs, educational institutions, and other organizations, and often include additional details such as household members, life events, and possessions, offering a rich context for studying elite community structure and interaction \cite{obrien2025gender,Broad1996social}. Each register serves as a comprehensive record of a local social network, defined according to the authors' notion of prominence. Despite widely recognized value of social register texts for over half a century \cite{obrien2024family,Broad2020social,Baltzell1958philadelphia}, they have been used sparingly because they are very laborious to collect and use for applied research. A vast amount of knowledge about elite communities has thus been missing from social science, despite the widely recognized importance of club membership to the study of elite communities \cite{bond2012bases,weber1947theory,mills1956power,keller1963beyond}.

The six registers are selected to support ongoing research into historical elite networks, drawing from several U.S. (and one non-U.S.) locations, over the late 19th to mid-20th centuries (Table~\ref{tab:affil-graph-details}).\footnote{Our dataset, drawn from six books, actually includes 18 separate text/network pairs, since the \textit{The Southern Social Register} \cite{whitmarsh1951} contains 13 individual sections for each state, which we separate.}
The texts themselves represent some of the variety in layout and writing styles used for social registers (Fig.~\ref{fig:pipeline}).

\vspace{0.3em}
\noindent\textbf{Bipartite affiliation graphs (Table~\ref{tab:affil-graph-details}).}
Each book yields a bipartite affiliation graph of
\textsc{Member} relational edges between
\textsc{Person} and \textsc{Club} entity nodes
\cite{breiger1947duality,lattanzi2009affiliation},
represented as a collection of
$\langle$\textsc{person}; \textsc{member}; \textsc{club}$\rangle$
triplets: for example, 
$\langle$``\textit{BELLASIS, John William Melville}'';
  \textsc{member}; 
  ``\textit{Byo.}''$\rangle$
in Fig.~\ref{fig:pipeline}'s \textit{Rhodesia} snippet.
Affiliation graphs are widely used to model memberships in elite groups, terrorist organizations, and other social networks, and are well-suited for studying diverse social science questions \citep{PREMSANKAR2015exploratory, tisch2023top,corradi2024,isah2015bipartite,holme2007korean,Marotta2014BankFirmCN,duxburyidentifying}.

\vspace{0.3em}
\noindent\textbf{Expanded knowledge graphs (Table~\ref{tab:expanded-kg-details}).}
For the Miami, Boston, and New York City datasets, we further expanded the KGs by including additional relation types, such as \textsc{College}, \textsc{Event}, \textsc{Plane Name}, and \textsc{Yacht Name}, reflecting the complexity commonly found in NLP research \cite{doddington-etal-2004-automatic,Riedel2010ModelingRA,zhang-etal-2017-position,han-etal-2018-fewrel,yao-etal-2019-docred,gardent-etal-2017-creating,huguet-cabot-etal-2023-red}. A full list of labeled relation types is provided \S\ref{app:textaffpairs}.

\vspace{0.3em}
\noindent\textbf{Construction of ground truth triplets.} 
To construct ground truth triplets, we developed a semi-automatic approach based on the observation that each book, despite containing relatively free-form text, followed several complicated patterns when presenting information. These patterns varied significantly across social register texts, requiring tailored strategies for each dataset. To capture the patterns, one author designed regular expressions and rule-based algorithms to extract triplets of $\langle\textsc{Person},\textsc{Member},\textsc{Club}\rangle$ relations for affiliation graphs, or of multiple relationship types for expanded KGs. For validation, two additional authors independently annotated four to six randomly selected pages from different sections of each book, resulting in approximately 200 ground truth triplets per book. To ensure consistency across all datasets, all authors first discussed and agreed on a standardized labeling scheme prior to annotation.

The performance of the regular expression and rule-based algorithm approach was evaluated by having the author responsible for its design to compare its output against the manually curated ground truth. This comparison confirmed that, with the exception of rare cases involving address relationships in the \textit{NYC} dataset (see \S\ref{app:ground-truth-construction}), the automated extractions achieved 100\% precision and recall. For the smaller social register text centering on  \textit{Denver}, where the total number of pages was limited, two authors manually annotated the entire dataset instead of using any regular expressions.

To enable the application of rule-based algorithms to the text, we first converted the scanned book images into machine-readable text using Claude Sonnet \cite{anthropicTheC3} as an OCR model. We then manually corrected transcription errors by systematically comparing the OCR output with the original scanned images to ensure accuracy.

\vspace{0.3em}
\noindent\textbf{Significant manual effort required.} The regular expressions and rule-based algorithms were highly complex, with up to $540$ lines of code, due to the variability and intricacy of the text formats. In the \textit{Southern US} states and \textit{Miami} datasets, entire families were described in contiguous blocks of text, using complex structural patterns based on titles to identify both people and their club memberships. In the \textit{Rhodesia} dataset, the varying formats for club memberships within text blocks required flexible regular expressions. For \textit{Boston} and \textit{New York City} books, the text formats lacked explicit family delineations. Family members were listed sequentially---adults first, then children, all on separate lines---with addresses, phone numbers, and events spanning multiple lines on the right side of the page. Club information for a person might appear on the same line as their name or on the following line next to another individual's name, linked correctly only through punctuation like brackets. 

Constructing these regular expressions and annotating parts of the ground truth demanded significant manual effort, in particular for the older social register texts. This labor-intensive process underscores the need for more efficient methods to reduce manual intervention. We propose using the high quality labels generated here as benchmarks to evaluate alternative approaches that require less manual effort.

\section{Benchmarking LLMs for Extracting Local and Global Structure}
\label{sec:findings}

\textsc{AffilKG} is the first ground truth resource to support both micro-level evaluation at the edge level---consistent with evaluations in the NLP literature---and macro-level evaluation based on high-level graph structure, which is highly relevant for real-world applications. In our preliminary evaluation, we assess performance at both levels, with our macro-level analysis specifically examining club membership sizes as a representative graph property. Given our goal of improving scalability, we focus our evaluation on low-resource methods, and use the following experimental pipeline:

\vspace{0.3em}
\noindent\textbf{Step 1: OCR.} \textit{Input: Image scans; Output: Text.} 
A vast amount of text, encompassing millions of historical documents, exists solely in print \cite{gotscharek2009lexical,piotrowski2012natural}, making OCR an essential step in extracting KGs from these texts. We compare two OCR models: Google DocumentAI\footnote{\url{https://aws.amazon.com/textract/}}, which outperforms prior methods \cite{Hegghammer2022ocr}, and Gemini-1.5-pro, which we find to also perform well.
\vspace{0.3em}

\noindent\textbf{Step 2: LLM.} \textit{Input: Text; Output: Relation triplets of $\langle$\textsc{person}; \textsc{member}; \textsc{club}$\rangle$}. To perform RE with LLMs as in existing literature, we use 1-shot in-context learning by providing an LLM with instructions about triplets to extract from an entry of text, plus one illustrative example (\S\ref{app:prompts}; Fig.~\ref{fig:pipeline}). We evaluate four LLMs: two proprietary models---Gemini-1.5-pro \cite{geminiteam2024geminifamilyhighlycapable} and GPT-4o \cite{openai2024gpt4ocard}---and two open-weight models, Llama 3 (8B and 70B) \cite{grattafiori2024llama3herdmodels}. Each model uses the same prompt.

\begin{table}
\centering
\fontsize{8.3pt}{8.5pt}\selectfont
\begin{tabular}{lllllll}

Dataset & OCR & {Metric} & Gem & 4o & ll-70b & ll-8b \\
\midrule
\multirow{4}{*}{Denver} 
 & \multirow{2}{*}{\fontsize{8pt}{8pt}\selectfont Doc AI} & {\fontsize{8pt}{8pt}\selectfont F1 (\%)} & 92.8 & 90.0 & 85.5 & 66.0 \\
 &                          & {\fontsize{8pt}{8pt}\selectfont RMAE} & 0.03 & 0.07 & 0.08 & 0.31 \\
 & \multirow{2}{*}{\fontsize{8pt}{8pt}\selectfont Gemini}  & {\fontsize{8pt}{8pt}\selectfont F1 (\%)} & 95.9 & 94.1 & 91.5 & 90.6 \\
 &                          & {\fontsize{8pt}{8pt}\selectfont RMAE} & 0.01 & 0.03 & 0.02 & 0.06 \\
\midrule
\multirow{4}{*}{Tenn.}
 & \multirow{2}{*}{\fontsize{8pt}{8pt}\selectfont Doc AI} & {\fontsize{8pt}{8pt}\selectfont F1 (\%)} & 93.8 & 91.4 & 84.2 & 57.1\\
 &                          & {\fontsize{8pt}{8pt}\selectfont RMAE} & 0.04 & 0.13 & 0.23 & 0.40 \\
 & \multirow{2}{*}{\fontsize{8pt}{8pt}\selectfont Gemini}  & {\fontsize{8pt}{8pt}\selectfont F1 (\%)} & 94.3 & 91.8 & 84.2 & 54.0 \\
 &                          & {\fontsize{8pt}{8pt}\selectfont RMAE} & 0.04 & 0.12 & 0.21 & 0.37 \\
\midrule
\multirow{4}{*}{Rhod.} 
 & \multirow{2}{*}{\fontsize{8pt}{8pt}\selectfont Doc AI} & {\fontsize{8pt}{8pt}\selectfont F1 (\%)} & 76.8 & 79.1 & 69.9 & 60.9 \\
 &                          & {\fontsize{8pt}{8pt}\selectfont RMAE} & 0.15 & 0.21 & 0.14 & 0.19 \\
 & \multirow{2}{*}{\fontsize{8pt}{8pt}\selectfont Gemini}  & {\fontsize{8pt}{8pt}\selectfont F1 (\%)} & 83.7 & 86.1 & 77.5 & 68.9 \\
 &                          & {\fontsize{8pt}{8pt}\selectfont RMAE} & 0.19 & 0.19 & 0.17 & 0.17 \\
\bottomrule
\end{tabular}
\caption{Edge-triplet F1 (\%) and club-size error (RMAE) for the top 10 largest clubs, across four LLMs on OCR outputs from better (Gemini) and worse (TextRact) OCR models.\label{tab:rmae-f1}}
\label{tab:results}
\end{table}

\vspace{0.3em}
\noindent\textbf{Step 3: Analysis of the extracted KG.} We perform the experiments over the text of \textit{Denver}, \textit{Tennessee}, and \textit{Rhodesia} in \textsc{AffilKG} and measure each model's output triplets using two metrics:

\vspace{0.3em}

\noindent\textit{\textbf{Triplet precision, recall, and F1 (Edge F1).}} In line with hundreds of previous papers \cite{zhao2024comprehensive,DETROJA2023survey}, we compute precision, recall, and F1 on relation triplets of $\langle$\textsc{person}; \textsc{member}; \textsc{club}$\rangle$ across $4$ LLMs, over texts outputted by the two OCR models. We refer to these as "edge correctness" metrics because each triplet uniquely defines an edge in the affiliation graph, where \textsc{person} and \textsc{club} entities correspond to nodes and \textsc{member} relations correspond to edges. A true positive requires that all three components of an LLM triplet---\textsc{person}, \textsc{member}, and \textsc{club}---match that of a ground truth triplet (details in \S\ref{app:eval}). 
Table~\ref{tab:rmae-f1} shares F1 performance across four LLMs on text produced by two OCR models, revealing substantial variability: differences between LLMs on the same OCR output range from $5$ to $40$ points, with the highest F1 reaching $95.9$.

\vspace{0.3em}

\noindent\textit{\textbf{Relative Mean Absolute Error (RMAE) for number of people in a club.}} Among many graph analyses that are useful for real applications, we select a simple one that captures higher-level KG structure: number of people in a club. Club membership is deemed to be vital to how elite communities socialize and cohere as a social group, as they convey status, facilitate socialization and the development of norms within elite communities \cite{Accominotti2018cultural,cousin2014globalizing}. To compute the number of people in a club $c$ on an affiliation graph, we count the number of neighbors each club node has, as: $n_c = \sum_{neighbors(c)} 1$. To quantify errors, we use the following error formula:

\vspace{-.4em}
{\footnotesize \[ \text{RMAE} = (1/|C|) \sum_{c \in C}   |\hat{n}_{c}-{n}_{c}| / n_{c} \] }
\vspace{-1em}

\noindent where $C$ is the gold-standard set of clubs, and $n_c$ and $\hat{n}_c$ are the true and predicted sizes of club $c$ respectively. Our analysis focuses on the ten largest clubs, as they typically play the most central roles in elite social networks---serving as hubs of influence, status, and social cohesion---and most strongly influence downstream applications. Details about these clubs are provided in \S\ref{app:10largest}. Table~\ref{tab:rmae-f1} shows that club-size RMAE may be as low as $0.01$ and as high as $0.37$.

\vspace{0.3em}
\noindent\textbf{Implications for future work.} Our preliminary results on three \textsc{AffilKG} datasets show that OCR and LLM combinations can construct meaningful KGs for real-world use, but performance varies significantly by model and dataset.  Higher edge accuracy corresponds to lower club-size errors in \textit{Denver} and \textit{Tennessee}, but not in \textit{Rhodesia}, suggesting other factors at play. Future research should test additional OCR models and LLMs, apply the approach to all \textsc{AffilKG} datasets, and investigate how LLM errors impact different types of graph analyses relevant to real-world applications.

\section{Conclusion}

\textsc{AffilKG} is the first dataset collection to pair complete scanned books with large, labeled affiliation and knowledge graphs. By providing high-quality annotations across diverse historical and geographical contexts, \textsc{AffilKG} enables researchers to investigate how errors in relation extraction propagate to downstream graph analyses---crucial for real-world applications---and supports the development of more robust KG extraction methods.

\section*{Limitations}

We did not carry out extensive prompt engineering for LLMs because our goal was to explore the impact of LLM errors on higher-level graph analyses, rather than to push performance. Early tests showed that rewording instructions produced similar results. Our prompts included detailed instructions and a carefully chosen example. 

While semi-structured text (i.e., text that may have a few patterns, such as names on the first line) may seem easier to extract affiliation graphs from, its prevalence in real applications \cite{obrien2024family,Broad2020social,Baltzell1958philadelphia} and the numerous errors by all LLMs on all texts in \textsc{AffilKG} highlight the need for its study. In particular, performance on \textit{Rhodesia} struggles significantly more overall.  

\section*{Ethical considerations}

The collected manual annotations (for the Denver book) were conducted by two of the coauthors. No human subject study was performed.

The collection of these social register books is part of a larger project to map out and analyze elite social networks, which may shed light on social questions about the dynamics of power and community, especially among individuals at the top of a social hierarchy. The books are historical, focusing on elite adults from 80 to 110 years ago, and do not uniquely identify living individuals.

While this paper's focus is on historical social networks, accurate inference of contemporary social networks has numerous uses not only in academic social science, but also applications with more complex ethical considerations such as marketing, law enforcement, etc.

\section*{Acknowledgments}

We thank anonymous reviewers, members of the UMass Natural Language Processing group, and participants in the UMass Computational Social Science Institute seminar, for feedback.
Thanks also to Sol Doeleman, Jia Sharma, Izaura Solipa and Shih-Yen Pan for additional comments and assistance.  This material is based in part upon work supported by National Science Foundation award 1845576. Any opinions, findings and conclusions or recommendations expressed in this material are those of the authors and do not necessarily reflect the views of the National Science
Foundation.

\bibliography{custom}

\appendix

\section{Appendix}
\label{sec:appendix}

\subsection{Dataset licensing}
All our new gold-standard dataset annotations will be released with a creative commons license, along with the open-source licensed codebase used for experiments.  To support high quality research, we also release scans of the books, some of which may be under copyright.

\subsection{Text and knowledge graph information}
\label{app:textaffpairs}

Table~\ref{tab:affil-graph-details} presented details about the text-affiliation graph pairs from \textit{Denver}, \textit{Rhodesia}, \textit{Miami}, \textit{Boston}, \textit{NYC}, and a single Southern US state (\textit{Tennessee}, which corresponds to the affiliation graph with the median number of person nodes across 13 southern US states). In Table~\ref{tab:metadata-more}, we provide details about text-affiliation graph pairs over all of the Southern US states.

Table~\ref{tab:expanded-kg-details} presented the number of relation types in the expanded knowledge graphs for social register texts from \textit{Miami}, \textit{Boston}, and \textit{NYC}. For the \textit{Miami} texts, the relation types include: Address; Address - Northern; Address - Summer; Address - Permanent; Address, Yacht; Clubs; College; School; Cruiser; Yacht; Plane; Sailer; and Racer. In the \textit{Boston} texts, the relation types are: Address, Club, Phone, College, and Event. For the \textit{NYC} texts, the relation types are: Address, Club, College, and Event.

\begin{table}
{\footnotesize\begin{tabular}{cccc}
 Loc./Year & \begin{tabular}{c} 
\# \\
pages
\end{tabular} & \begin{tabular}{c} 
\#~nodes \\
(org/person)
\end{tabular} & \begin{tabular}{c} 
\# \\
edges
\end{tabular} \\
\midrule 
NYC/1892 & 299 &  178 / 5553 & 14061 \\
Boston/1903 & 150 &  188 / 2534 & 6228\\
Denver/1931 & 34 & 248/358 & 1352 \\
Alab./1951 & 49& 236/401&753 \\
Ark./1951 & 22& 83/167 & 292 \\
Flor./1951 & 164& 728/1395 & 3208 \\
Geor./1951 & 100& 289/685 & 1275 \\
Ken./1951 & 28& 121/227 & 501 \\
Lou./1951 & 67& 206/544 & 1328 \\
Miss./1951 & 26& 98/149 & 293 \\
N. Car./1951 & 51& 224/375 & 747 \\
S. Car./1951 & 33& 145/235 & 515 \\
Tenn./1951 & 47& 174/399 & 872 \\
Tex./1951 & 153& 394/993 & 2026 \\
Vir./1951 & 55& 274/401 & 995 \\
W. Vir./1951 & 31& 135/217 & 457 \\
Rhod./1963 & 169& 767/898 & 2367 \\
Miami/1965 & 503 &  1471 / 3775 & 11830 \\
\hline
\end{tabular}
}

\caption{Details about New York City, Boston, Denver, Southern US states, Rhodesia, and Miami text and affiliation graph pairs.}
\label{tab:metadata-more}
\end{table}

\subsection{Top 10 largest organizations}
\label{app:10largest}
The top 10 clubs as referenced in text, their full name, and number of elite members in \textit{Denver}, \textit{Tennessee}, and \textit{Rhodesia} are in Tables~\ref{tab:Denver-clubs}, \ref{tab:tn1951-clubs}, and \ref{tab:rhodesia-clubs} respectively.
\begin{table*}[h!]
\centering{\small
\begin{tabular}{llp{6.5cm}c}

Club name in text & Full club name & Description & \# members\\
\hline
D.\ C.\ C. & Denver Country Club & An exclusive organization devoted principally to golf and outdoor recreation. & 174\\
\midrule D.\ C. & Denver Club & One of Denver's most exclusive social organizations. Membership composed of financiers, representative business and professional men. & 150\\
\midrule M.\ C.\ of C. & Motor Club of Colorado & Devoted to the motoring interest. Membership composed of auto owners. & 110\\
\midrule D.\ A.\ C. & Denver Athletic Club & Devoted to athletics. Membership composed of prominent business and professional men. & 104\\
\midrule C.\ H.\ C. & Cherry Hill Club & A country club devoted to outdoor recreation. Membership composed of representative men and women. & 98\\
\midrule U.\ C. & University Club & An exclusive organization composed of representative business and professional men who are college graduates. & 64\\
\midrule M.\ H.\ C. & Mile High Club & An exclusive organization to entertain national and international notables at dinner. & 43\\
\midrule L.\ C.\ C. & Lakewood Country Club & A representative organization devoted to golf and outdoor recreation. & 42\\
\midrule R.\ C. & Rotary Club & An organization composed of prominent merchants from every line of trade. Formed for commercial and civic betterment. & 36\\
\midrule C.\ C. & Cactus Club & An organization for literary and dramatic purposes. Largely used as a downtown lunch and dinner club by professional and business men. & 35\\
\hline
\end{tabular}}
\caption{10 largest clubs in \textit{Denver} (1931), with descriptions and membership counts, where descriptions are copied from the Denver 1931 Social Register \citet{1931denver}.}
\label{tab:Denver-clubs}
\end{table*}

\begin{table}[h!]
\centering{\small
\begin{tabular}{p{3cm} c}

Club name in text & \# members \\
\hline
Junior League & 93 \\
\midrule
Fairyland & 73 \\
\midrule
Belle Mead Country & 65 \\
\midrule
Memphis Country & 63 \\
\midrule
Mountain City & 38 \\
\midrule
Cumberland & 26 \\
\midrule
Centennial & 25 \\
\midrule
Cotillion & 24 \\
\midrule
Daughters of the American Revolution & 23 \\
\midrule
Colonial Dames & 22 \\
\hline
\end{tabular}}
\caption{10 largest clubs in \textit{Tennessee} (1951) with membership counts.}
\label{tab:tn1951-clubs}
\end{table}

\begin{table}[h!]
\centering
{\small
\begin{tabular}{p{1.8cm} p{3cm} c}

Club name in text & Full club name & \# members\\
\hline
Sby. & Salisbury Club & 299 \\
\midrule
Byo. & Bulawayo Club & 146 \\
\midrule
Royals Byo. Golf & Royal Bulawayo Golf Club & 88 \\
\midrule
Byo. Country & Bulawayo Country Club & 55 \\
\midrule
Ruwa Country & Ruwa Country Club & 51 \\
\midrule
Sby. Sports & Salisbury Sports Club & 48 \\
\midrule
Umtali & Umtali Club & 44 \\
\midrule
New & New Club & 42 \\
\midrule
Lon. & London Club & 35 \\
\midrule
Rotary & Rotary Club & 30 \\
\hline
\end{tabular}}
\caption{10 largest clubs in \textit{Rhodesia, Central and East Africa} (1963) with abbreviated/full names and membership counts.}
\label{tab:rhodesia-clubs}
\end{table}

\subsection{Details for constructing regular expressions to extract ground truth triplets}
\label{app:ground-truth-construction}

\textbf{Affiliations.} For the \textit{Southern US} and \textit{Miami} texts, the rules involved identifying the string “Clubs:” and checking for titles, such as “Mr.,” “Mrs.,” “Dr.,” “Capt.,” “Lieut.” If a title was present, the associated clubs were linked to the adult that had the same title; otherwise, all clubs were associated with all adults in the text entry, referred to on the first line. For the \textit{Rhodesia} text, the rules consisted of checking for the "Club/s:" string, and splitting the content afterward by commas, semicolons, and "and" delimiters, until the next listed category indicated by a colon; e.g., "Rec/s:", "Add:" --- the text was not consistent on the delimiter that it used and whether it used a plural. For the \textit{Boston} and \textit{NYC} texts, abbreviations delimited by periods were identified as possible clubs. Clubs were linked to the individual listed on the same line (the male if two individuals appeared), or, if listed after a bracket, to the person named on the line above.

\textbf{Other relations.} In the \textit{Boston} and \textit{NYC} datasets, addresses can span multiple lines on the right side of the page, across household members. Household boundaries are indicated either by a new last name or a vertical line on the page, but the vertical line is often missed by OCR. Therefore, our rule-based algorithms distinguish households as follows: (1) Each new last name indicates a new household. (2) If the last name matches the one on the previous line, a new household is created if both individuals are adults and the previous address is complete. Adulthood is identified by titles such as "Mrs.," "Dr.," or "Prof."---distinguishing whether "Mr." signifies an adult or child is more ambiguous. To address this, a further rule specifies: if both the previous line and current line addresses are complete, the individual on the current line is in a different household from that on the previous line---this is consistent with the assumption that children rarely have their own address. (3) If a last name is missing, the individual is not in a new household.

Persons are identified using complex regular expressions due to varied name formats. Clubs are captured as any 1–4 character string followed by a period or space-period. Colleges are identified by strings with an apostrophe and year (e.g., "Harv'92"). Phone numbers are identified by strings that start with "Phone" or variations (e.g., "Phone No", "Ph no") and have a number after. Because events and addresses often span multiple lines, the algorithm processes all lines for each household, removing any segments containing person or club names. The rest of the text is parsed as events and addresses: events are indicated by the presence of a calendar month (e.g., ``Jly" for July in the specific book format), or the word "Married". All other non-matching text is treated as address information. If multiple adults appear on the same line, only male adults are linked to clubs and colleges. However, all household members are linked to addresses, events, and phone numbers.

For the \textit{Miami} dataset, relationships are extracted by searching for the string of the relationship type. For relationships other than clubs, colleges, or schools, each adult in the household is linked to the relation. For clubs, colleges, or schools, titles such as "Mr.," "Mrs.," or "Dr." are checked: if a title is present, associated clubs, colleges, or schools are linked to the corresponding adult with that title; otherwise, all such relations are associated with every adult referred to in the first line of the text entry.

\subsection{LLM prompts} 
\label{app:prompts}
We used the same prompts across LLMs for each dataset. The prompt for the Denver dataset was:

\begin{quote}
\textit{[text entry]}

\textit{Extract relations in the format of person-name; member; club from the text, indicating that person-name is a member of club. In the text, — indicates a different club. List each relation as a bullet point. An example is: }

\textit{[example text entry and outputted relation triplets]}
\end{quote}

The prompt for the Southern US dataset was:

\begin{quote}
\textit{[text entry]}

\textit{Extract relations in the format of person-name; member; club from the text, indicating that person-name is a member of club. Schools are not clubs. If no title such as 'Mr.', 'Mrs.', 'Miss', 'Rev.', 'Dr.', etc. is in front of a list of clubs, then the first two persons mentioned are members in each of the listed clubs. Otherwise, only the person with the corresponding title is in each of the listed clubs. List each relation as a bullet point.}

\textit{[example text entry and outputted relation triplets]}
\end{quote}

The prompt for the Rhodesia dataset was: 

\begin{quote}
\textit{[text entry]}

\textit{Extract relations in the format of person-name; member; club from the text, indicating that person-name is a member of club. Consider only the clubs that are listed after ``Club(s):". List each relation as a bullet point. An example is:}

\textit{[example text entry and outputted relation triplets]}
\end{quote}

\subsection{Details for evaluating triplets}
\label{app:eval}

For evaluation, we compared the triplets $\langle$\textsc{Person}, \textsc{member}, \textsc{Club}$\rangle$ outputted by the LLM to those in the ground truth set. A true positive required that all three components of an LLM triplet match that of a ground truth triplet. However, we allowed more flexible matching for individual \textsc{Person} and \textsc{Club} entities. 

We allowed this flexibility because we observed that LLMs sometimes omit middle initials, names, or suffixes (e.g., Jr., Sr., II) in \textsc{person} names. For \textsc{club} names, LLMs sometimes included additional information, such as relevant parenthetical details, truncated a very long name, or expanded abbreviations. We did not penalize the LLM in these cases because these were spelling, rather than semantic, issues. 

Therefore, we identified a positive entity match when either the LLM entity matched the ground truth exactly, or when both exceeded six characters and one was a substring of the other. We also allowed matches between abbreviations and full names (e.g., "Bulawayo" and "Byo"; "Assn" and "Association"). All spaces and punctuation were removed during the comparison. For \textsc{Person} entities, we also required that titles such as "Mr." or "Mrs." needed to be correct if present in the text. We validated this approach by manually checking the results of 150 entity pair comparisons for each dataset.

\end{document}